# MIDV-500: a dataset for identity document analysis and recognition on mobile devices in video stream

*V.V. Arlazarov* [1,2,3], *K. Bulatov* [1,2,3], *T. Chernov* [3], *V.L. Arlazarov* [1,2,3]
[1] *Moscow Institute of Physics and Technology (State University), Moscow, Russia,*
[2] *Institute for Systems Analysis, FRC CSC RAS, Moscow, Russia,*
[3] *LLC "Smart Engines Service", Moscow, Russia*

*Abstract*

A lot of research has been devoted to identity documents analysis and recognition on mobile devices. However, no publicly available datasets designed for this particular problem currently exist. There are a few datasets which are useful for associated subtasks but in order to facilitate a more comprehensive scientific and technical approach to identity document recognition more specialized datasets are required. In this paper we present a Mobile Identity Document Video dataset (MIDV-500) consisting of 500 video clips for 50 different identity document types with ground truth which allows to perform research in a wide scope of document analysis problems. The paper presents characteristics of the dataset and evaluation results for existing methods of face detection, text line recognition, and document fields data extraction. Since an important feature of identity documents is their sensitiveness as they contain personal data, all source document images used in MIDV-500 are either in public domain or distributed under public copyright licenses.

The main goal of this paper is to present a dataset. However, in addition and as a baseline, we present evaluation results for existing methods for face detection, text line recognition, and document data extraction, using the presented dataset.

<u>Keywords</u>: document analysis and recognition, dataset, identity documents, video stream recognition.

<u>Citation</u>: Arlazarov VV, Bulatov K, Chernov T, Arlazarov VL. MIDV-500: a dataset for identity document analysis and recognition on mobile devices in video stream. Computer Optics 2019, 43(5): 818-824. DOI: 10.18287/2412-6179-2019-43-5-818-824.

<u>Acknowledgements</u>: This work is partially supported by Russian Foundation for Basic Research (projects 17-29-03170 and 17-29-03370). All source images for MIDV-500 dataset are obtained from Wikimedia Commons. Author attributions for each source images are listed in the description table at *ftp://smartengines.com/midv-500/documents.pdf* .

## Introduction

Smart phones and mobile devices have become the de-facto way of receiving various government and commercial services, including but not limited to e-government, fintech, banking and sharing economy [1]. Most of them require the input of user's personal data [2]. Unfortunately, entering data via mobile phone is inconvenient, time consuming and error-prone. Therefore many organizations involved in these areas decide to utilize identity document analysis systems in order to improve data input processes. The goals of such systems are: to perform document data recognition and extraction, to prevent identity fraud by detecting document forgery or by checking whether the document is genuine and real, and others. Modern systems of such kind are often expected to operate in unconstrained environments. As usual, it is essential for research and development of these systems and associated methods to have access to relevant datasets.

However, identity documents are special in a sense that they contain sensitive personal information. This creates complications in several aspects. First, storing any personal data presents a security risk in case it is leaked, resulting in identity fraud and significant financial damages both for corresponding ID holders and the party responsible for leaking the data. Second, people understand that risk and are not as easily convinced to share their personal data with someone they do not trust. Third, identity documents are uniquely bound to their owners which makes them very rare, increasing the costs of data collection even further. Fourth, there are very few publicly available samples of each identity document and most of them are protected by copyright laws which makes them unusable in research. Finally, in many countries it is illegal not only to distribute personal data but even collect and store it without a special permission [3].

For the reasons explained above, there currently are no publicly available datasets containing identity documents. Research teams are forced to create and maintain their own datasets using their own resources. This can be easily observed by looking through the recent papers devoted to identity document analysis, for instance [4–6]. For some teams having access to such sensitive data is not a problem, especially if they are industry-oriented, already have a product related to identity document analysis, have government support and possess the required expertise in data security. Commercial companies working in this field might receive large datasets from their customers, but they will never share it because of aforementioned legal, security and also market advantage reasons. However, for most researchers these options are not easily obtainable.

The absence of publicly available identity document datasets is a serious problem which directly affects the state of research in this field. It raises the entry barrier, putting off those who do not have the resources to create their own datasets and slowing down those who have not collected the data yet. Furthermore, it becomes impossible to evaluate and compare various identity document analysis methods to each other, since they have been tested on completely different and locked down data [7, 8]. In addition, there is an ethical concern regarding verifiable and reproducible research, especially during peer review process. Even great papers presenting state-of-the-art results can be rejected by the scientific community for being impossible to verify without the testing data.

Some classes of publicly available datasets may be useful for researching common subtasks of identity document analysis and recognition. Examples of such tasks and relevant datasets are listed in Table 1. For the scope of problems related to a more general identity document analysis pipeline, however, these datasets prove insufficient, especially considering mobile video stream recognition systems.

*Table 1. Related tasks and public datasets*

| Task | Datasets |
|---|---|
| Document detection and localization | [9] |
| Text segmentation | [10] |
| Document image binarization | [11] |
| Optical character recognition | [12, 13] |
| Image super-resolution | [14] |
| Document forensics | [15] |
| Document image classification | [16] |
| Document layout analysis | [17, 18] |
| Document image quality assessment | [19, 20] |
| General image understanding | [21, 22] |

In this paper we present a Mobile Identity Document Video dataset (MIDV-500), which in contrast to other relevant publicly available datasets can be used to develop, demonstrate and benchmark a coherent processing pipeline of identity document analysis and recognition in its modern applications and use cases. All source document images used in MIDV-500 are either in public domain or distributed under public copyright licenses which allows to use these images with attribution to the original publisher. The dataset is available for download at *ftp://smartengines.com/midv-500/* .

### *1. Problem statement and use case*

#### *Identity document properties*

When developing a system for identity document recognition it is important to take into account special properties of identity documents. As mentioned earlier, identity documents hold sensitive personal data which also makes the cost of recognition errors quite high compared to common text recognition tasks. In order to provide additional level of forgery protection identity documents often have complex graphical background possibly containing guilloche, watermarks, retroreflective coating which is prone to glare, holographic security elements which change their appearance depending on relative positions of camera and document [23]. Moreover, text is printed with special fonts, sometimes using indent or embossed printing techniques.

Another feature of identity documents is their fields layout, which is often strict in terms of absolute and relative position and size of document elements. However, there could be complex exceptions, such as optional text fields which may or may not be present on particular document, floating textual and graphical elements, etc.

#### *Uncontrolled recognition environment conditions*

Modern mobile device recognition use cases usually involve unknown and uncontrolled environment and a person who handles the recognition system often is not familiar with how such system operates. Uncontrolled or unconstrained environment means that scene geometry, lighting conditions and relative movement model of the object and the capturing device are unknown. It also implies that there is no guarantee for the input data to be of high quality and indicates the possible risk of information loss.

Depending on which particular frame has been captured, the document data information might be partially or completely lost because of document not being fully present inside the frame, camera not being fully focused, bad or uneven lighting conditions, glares, camera noise and other common distortions [24].

A possible approach which could be used to solve the problems related to uncontrolled capturing conditions is the analysis of a video stream instead of a single image.

#### *Video stream recognition*

One of the approaches to increase the precision of information extraction is recognition in a video stream [14, 25]. Despite having to process larger amounts of graphical data, usage of video stream allows to employ methods inaccessible when analyzing a single image:

1. Using multiple input images of the same object it is possible to employ the "super-resolution" techniques [14] for obtaining a higher quality image, suitable for further processing and high-precision data extraction.
2. In scope of the tasks of object detection and localization the usage of video stream allows to consider the tasks of filtering and perform refinement using the processing result from the previous frames [26].
3. Having a sequence of video frames containing the same object a problem statement can be formulated for methods of selecting the "best" suitable object image, from the perspective of image quality [24] and suitability for certain methods of image processing and information retrieval [27].
4. When solving the task of object recognition or classification having multiple image frames it is possible to perform recognition of the object in each frame and combine the recognition results thus increasing the expected precision. Moreover, using video stream analysis a recognition system could be constructed according to the principles of Anytime algorithms [28], such that it is possible to regulate the trade-off between expected precision and total processing time, by varying the stopping rule parameters.

Recognition in a video stream has a high potential as an approach to the task of automatic document data extraction, especially considering the mobile device recognition systems. Thus, creation and maintenance of substantial datasets on this topic is of high importance.

### *The goals of this paper*

Authors would like to point out that the goal of this paper is not to propose new methods of image or video stream recognition. The goals of this paper are the following:

1. To present a dataset. The description of the dataset is given in section 2.
2. To show that having a substantial dataset of images and videos of identity documents is important. We hope that it was shown in the introductory section.
3. To show that using a video stream expands the capabilities of recognition systems, as we demonstrated in this section.
4. To give examples of useful applications of the presented dataset for performing comparative and quantitative studies of identity document image processing methods. These will be given in section 3.

Authors hope that the dataset presented in this paper will help the computer vision, image processing, and document analysis and recognition communities to conduct high-quality, extensive, and reproducible research related to the following scientific challenges:

– Document detection and localization. For each frame in the dataset we prepared a ground truth with document boundaries quadrangle.
– Document identification. Dataset consists of video clips of various document types.
– Document layout analysis. Ground truth contains boundary coordinates of each document field.
– Face detection. For each unique document ground truth includes personal photo position.
– Optical character recognition. Ground truth contains ideal text values of each document field.
– Image quality assessment. Video clips present in the dataset may have glares, complex lighting, defocus, occlusions and other distortions.
– Automatic rejection of potentially incorrect recognition results.
– Engineering the recognition stopping decision rule in order to optimize the scanning process duration.
– Vanishing points detection based on document's edges and contents.
– The task of selecting the best image from a video sequence.
– Document image super-resolution and reconstruction. For each unique document a source image with minimal distortions is included in the dataset.

### *2. Dataset structure*

### *Image data description*

The MIDV-500 dataset contains video clips of 50 different identity document types, including 17 types of ID cards, 14 types of passports, 13 types of driving licenses and 6 other identity documents of various countries. Each document image was printed on a photo paper using HP Color LaserJet CP2025 printer and laminated using glossy film. Then for each document printout we have recorded video clips in 5 different conditions using two mobile devices (10 videos per document in total). Conditions are described in Table 2.

First letter of each identifier specifies the condition and the second letter specifies the mobile device on which the video clip was recorded. The two mobile devices used for recording were: Apple iPhone 5 and Samsung Galaxy S3 (GT-I9300). Examples of frames with each condition are presented on Fig. 1.

*Table 2. Dataset structure*

| Identifier | Description |
|---|---|
| TS, TA | "Table" – simplest case, the document lays on the table with homogeneous background |
| KS, KA | "Keyboard" – the document lays on various keyboards, making it harder to utilize straightforward edge detection techniques |
| HS, HA | "Hand" – the document is held in hand |
| PS, PA | "Partial" – on some frames the document is partially or completely hidden off-screen |
| CS, CA | "Clutter" – scene and background are intentionally stuffed with many unrelated objects |

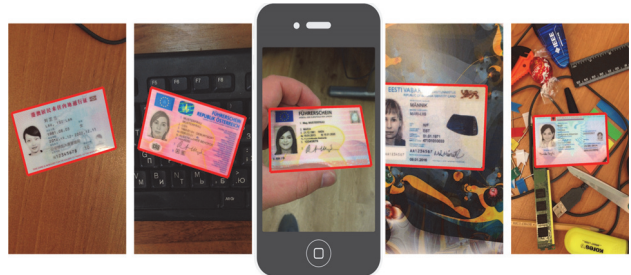

*Fig. 1. Example frames from MIDV-500 dataset.*
*5 different conditions are presented (left to right):*
*Table, Keyboard, Hand, Partial and Clutter*

Thus, 500 videos were produced: (50 documents)×(5 conditions)×(2 devices). Each video were at least 3 seconds in duration and the first 3 seconds of each video were split with 10 frames per second. In total there are 15000 annotated frame images in the dataset. Each captured frame had the same resolution of 1080×1920 px. Original images of each document type are also provided within the dataset for reference.

It should be noted that real identity documents usually have holographic security layers and other optical variable devices. The captured document samples for MIDV-500 dataset do not have those elements, hence for fraud detection and authentication tasks the dataset can only be used as a negative sample. In the interest of future research on such topics as the influence of lighting-related noise on identity documents analysis and recognition methods, more conditions might be considered the in scope of future work. Due to limited set of presented conditions and document types, using this dataset for training purposes could lead to possible overfitting. Thus, the main purpose of the MIDV-500 dataset is facilitate testing, evaluation, and comparison of document analysis methods.

*Ground truth*

For each extracted video frame we have prepared a document location ground truth (Fig. 2). The annotation was performed manually. If corners of the document are not visible on the frame, the corresponding coordinates point is extrapolated outside the frame (if the document is not visible on the frame at all, all four points will lay outside the frame boundaries).

```
{
  "quad": [ [0, 0],      [111, 0],
            [111, 222], [0, 222] ]
}
```

*Fig. 2. Frame ground truth JSON format*

For each unique document we have prepared ground truth containing its geometric layout represented by document field quadrangles, and UTF-8 string values for text fields. The ground truth is given in JSON format (Fig. 3).

```
{
  "field01": {
    "value": "Erika",
    "quad": [ [983, 450],   [1328, 450],
              [1328, 533], [983, 533] ]
  },
  // ...
  "photo": {
    "quad": [ [78, 512],    [833, 512],
              [833, 1448], [78, 1448] ]
  }
}
```

*Fig. 3. Unique document ground truth JSON format*

In total in the dataset there are 48 photo fields, 40 signature fields and 546 text fields. In addition to fields written using Latin characters with diacritics the dataset contains fields written using Cyrillic, Greek, Chinese, Japanese, Arabic and Persian alphabets.

### 3. Experimental baselines

The main goal of this paper is to present an open dataset of videos containing identity documents captured with mobile devices. Authors believe that this dataset's structure allows it to be used for evaluating various computer vision and document analysis methods (see section 1).

In order to perform quantitative experiments in relation to computer vision methods the dataset should contain ground truth on multiple levels. For example, in order to estimate the effectiveness of document detection and type classification methods it should have ideal coordinates of document boundaries on each frame and document type labels; for per-field segmentation it is necessary to have ideal field coordinates, etc. In order to assess the applicability of MIDV-500 for conducting such studies we conducted two comparative experimental evaluations:

1. Evaluation of face detection methods;
2. Evaluation of text string recognition methods.

Combined annotation of document boundaries for each frame and of objects coordinates in the document layout allows to perform not only studies of object detection and information extraction algorithms and measure their accuracy characteristics, but also evaluate their robustness against various distortions. The latter is especially relevant for performing objective studies of complex document analysis systems, where the result of each processing stage is predicated by the accuracy of its predecessor in the workflow. Thus in this section not only the accuracy of the face detection and text recognition methods is measured, but also their robustness is studies using a noise modeling.

Besides methods for solving isolated tasks, as document recognition implies complex analysis of the image with extraction of all document fields, at the end of this section we give such results, obtained using an existing document recognition system Smart IDReader [25].

The results of comparative experiments performed in this section serve as an example of using MIDV-500 and provide a baseline for future work based on the dataset.

*Face detection*

For face detection baseline we have considered only a part of the dataset according to the following criteria. First, the face must be present on the document, which is equivalent to "photo" field presence in the document ground truth. Second, for each given frame, the document must be fully within the camera frame which can be verified by checking the "quad" ground truth coordinates against the frame size. This way, 9295 frames out of 15000 were filtered for the experiment.

Face detection was performed using open source libraries dlib 18.18 [29] and OpenCV 3.4.3 [30] with default frontal face detectors. It was conducted in two modes: using original frames, and using projectively restored document images based on ground truth document coordinates. Default parameters for face detection were used.

In order to account for different physical sizes of the documents, the projective rectification was performed with 300 DPI resolution. To achieve this, for the sake of simplicity, the width of all card-size documents was assumed to be 86 mm, all passports – 125 mm and the only TD2-size ID card – 105 mm. The height was calculated proportionally, according to the aspect ratio of the document's source image.

The "photo" ground truth quadrangle was projectively transformed from the template coordinates to frame coordinates (according to the ground truth document boundaries), the same was done for detection results in the cropped document detection mode. We then used a relaxed binary evaluation metric for each frame – the face detection was considered correct if more than 50% of the largest found face rectangle laid inside the "photo" ground truth quadrangle. The detection quality was computed as the rate of correct face detections. Table 3 shows face detection results according to this metric.

*Table 3. Face detection evaluation results*

| Mode | dlib | OpenCV |
|---|---|---|
| Full frames | 87.53% | 76.41% |
| Cropped documents | 89.77% | 87.51% |

To evaluate the robustness of the face detection algorithms in cropped documents mode against the errors of document boundaries detection, the following coordinate noise modelling has been performed. Bivariate normally distributed noise has been added independently to each vertex $\mathbf{P}_i$ of the document boundaries quadrangle from ground truth:

$$\mathbf{P}_i \leftarrow \mathbf{P}_i + \mathbf{X}_i; \quad \mathbf{X}_i \sim N_2(\mathbf{0}, \sigma^2 \mathbf{I}_2); \quad \forall i \in \{1,2,3,4\}, \quad (1)$$

where $N_2$ denotes a normal distribution, $\mathbf{I}_2$ is a two-dimensional identity matrix, and $\sigma$ is a standard deviation of the noise distribution. As all frames in the dataset have the same resolution, for the sake of simplicity $\sigma$ will be denoted in pixels.

With the described noise applied to the document boundaries, projective restoration of document images was performed as before, for a range of $\sigma$ values (see example in Fig. 4). Then, the face detection algorithms were re-evaluated using the metric described above.

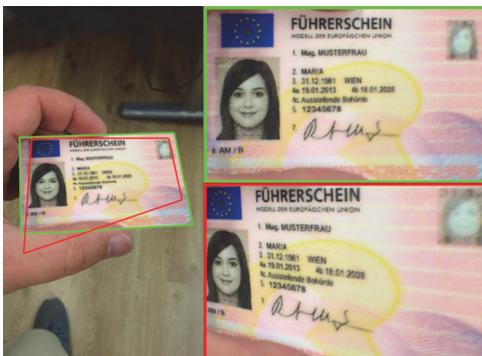

*Fig. 4. Projectively restored document images for the ground truth quadrangle (green, σ=0 px) and after applying coordinate noise (red, σ=50 px)*

Figure 5 demonstrates the face detection results with noisy document image rectification.

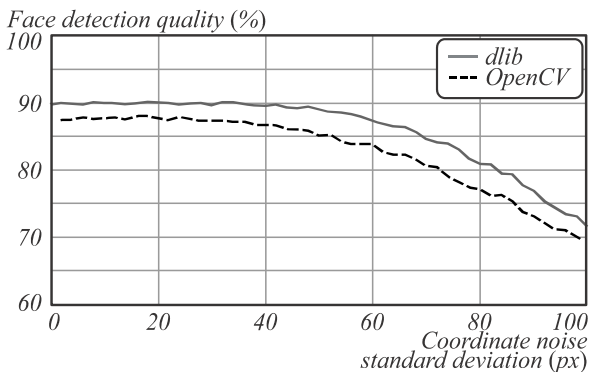

*Fig. 5. Face detection results with noisy prior document image restoration*

It can be observed that both frontal face detectors provide better quality when faces are detected in rectified document images compared to the original frames. Both detectors showed tolerance for prior document boundaries detection error with $\sigma < 40$ px given input frames resolution of 1080×1920 px.

### Text fields OCR

For text fields recognition baseline we have separately analyzed four identity document field groups: document numbers, numeric dates, name components (which contained only Latin characters without diacritics) and machine-readable zone (MRZ) lines. As in the previous experiment, only the frames on which the document is fully visible were considered. As an input to the OCR subsystem, each field was cropped according to the combined ground truth of document boundaries coordinates and template text field coordinates, with additional margins on each side equal to 10% of the smallest dimension of the text field. As in the previous experiment, each text field cropping size corresponded to 300 DPI resolution.

The recognition was performed using open-source libraries Tesseract 3.05.01 [31] and GNU Ocrad 0.26 [32] using default parameters for English language. A relaxed evaluation metric in form of a Normalized Levenshtein Distance [33] was used to demonstrate the relative amount of OCR errors:

$$V(r,w) = \frac{2 \cdot \text{levenshtein}(r,w)}{|r| + |w| + \text{levenshtein}(r,w)}, \quad (2)$$

where $r$ is the recognition result string, and $w$ is the correct text field value provided in the ground truth.

The character comparison was case-insensitive and latin letter O was treated as equal to the digit 0. Table 4 summarizes the obtained results.

*Table 4. Text field OCR evaluation results*

| Field group | Unique fields | Images in total | V(r,w) | |
|---|---|---|---|---|
| | | | Tesseract | Ocrad |
| Num. dates | 91 | 17735 | 0.3598 | 0.5856 |
| Doc. num. | 48 | 9329 | 0.4222 | 0.6769 |
| MRZ lines | 30 | 5096 | 0.2576 | 0.5359 |
| Latin names | 79 | 15587 | 0.4433 | 0.5962 |

In order to evaluate the robustness of the text extraction algorithms we used the same technique as with face detection experiment, using document boundaries detection noise model (1). Fig. 6 demonstrates the recognition results under noisy prior restoration of the document images from each frame.

As it can be seen from the graphs on Fig. 6, text recognition accuracy degrades even with small distortions, and even with padding applied to the field bounding boxes. Using the combined ground truth provided in MIDV-500 dataset it is possible and important to analyze document analysis stages not only in terms of their accuracy, but also their robustness against the errors of pre-processing stages of the document recognition pipeline.

### Full document processing

The aim of the last series of experiments was to evaluate a full document recognition system's accuracy on original MIDV-500 video clips, which are also provided along with extracted frames. For this purpose, a commercial video stream recognition system Smart IDReader 2.1.0 was used that is described in detail in [25].

The recognition was performed using "out of the box" Smart IDReader configuration without additional parameter tuning. Each of 500 video clips was passed to the system as an ordered series of frames and then the final recognition result was taken for evaluation. Moreover, for each of the video clips we specified a document type mask (country and ID document class) corresponding to the given clip as required by the system interface.

The accuracy metric was set to be an end-to-end recognition quality which is much stricter than the one used in previous experiments: the text field recognition result was counted as correct only if its value exactly matched the ground truth (with the only preprocessing of dates format, as the analyzed system date output format is standardized). Any field that had not been returned by the system was counted as incorrect. The final accuracy was measured as the ratio of correctly recognized fields.

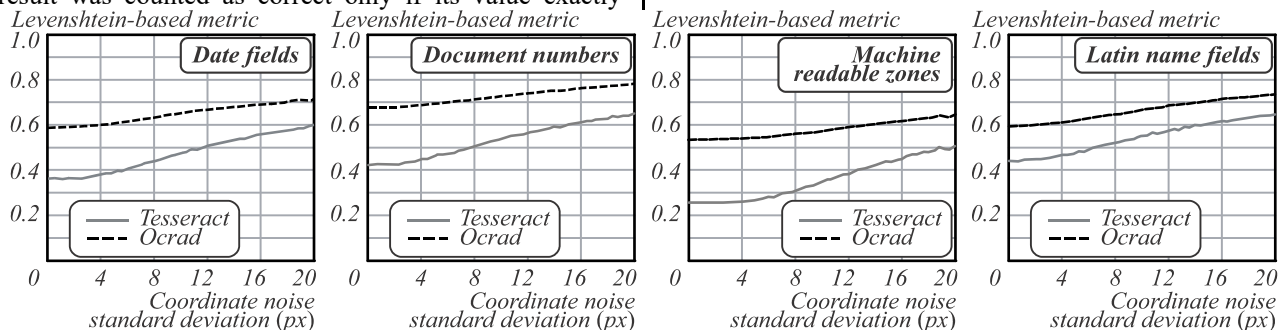

Fig. 6. Text field recognition results with noisy prior document image restoration

Table 5 shows the Smart IDReader recognition accuracy according to the aforementioned metric broken down by the capturing conditions (see Table 2 for more information about dataset structure) and for the whole dataset. All the fields were grouped into the same four categories as in the previous experiment (numeric dates, document numbers, MRZ lines and Latin names).

Table 5. Full document OCR evaluation results for various capturing conditions using Smart IDReader video stream document recognition system

| Field group | Field recognition accuracy (%) | | | | | |
|---|---|---|---|---|---|---|
| | TS, TA | KS, KA | HS, HA | PS, PA | CS, CA | All |
| Num. dates | 89.56 | 88.46 | 93.96 | 91.21 | 87.36 | 90.11 |
| Doc. num. | 88.54 | 77.08 | 82.29 | 72.92 | 77.08 | 79.58 |
| MRZ lines | 88.33 | 83.33 | 68.33 | 61.67 | 73.33 | 75.00 |
| Lat. names | 78.48 | 70.89 | 81.65 | 77.85 | 74.68 | 76.71 |

These are the first published recognition accuracy results for Smart IDReader on an openly accessible dataset. The authors hope that this will serve as a baseline for further development of identity document analysis systems.

## Conclusion

The primary goal of this paper was to present a dataset which is sufficient for conduction research in the field of automatic data extraction from identity documents. The MIDV-500 dataset was presented, containing 500 video clips of 50 different identity document types. To the best of our knowledge this is the first publicly available dataset for identity document analysis and recognition in video stream. Additionally the paper presents three experimental baselines obtained using the dataset: face detection accuracy, separate text fields OCR precision for four major identity document field types, and identity document data extraction from video clips. For the two former results we presented an analysis of robustness against document boundaries localization errors.

We believe this dataset will be a valuable resource for research in identity document analysis and recognition and will motivate future work in this field.

*Authors' information*

**Vladimir Viktorovich Arlazarov** (b. 1976), PhD, graduated from Moscow Institute of Steel and Allows in 1999, majoring in Applied Mathematics. Currently he works as head of division 93 at the Institute for Systems Analysis FRC CSC RAS. Research interests are pattern recognition and machine learning. E-mail: *vva@smartengines.com* .

**Konstantin Bulatovich Bulatov** (b. 1991) graduated from National University of Science and Technology "MISiS" in 2013, majoring in Applied Mathematics. Currently he works as a programmer (1st category) at the Institute for Systems Analysis FRC CSC RAS. Research interests are pattern recognition, combinatorial algorithms and computer vision. E-mail: *kbulatov@smartengines.com* .

**Timofey Sergeevich Chernov** (b. 1992), PhD, graduated from National University of Science and Technology "MISiS" in 2013, majoring in Applied Mathematics. Currently he works for Smart Engines Limited. Research interests are pattern recognition and computer vision. E-mail: *chernov.tim@smartengines.com* .

**Vladimir Lvovich Arlazarov** (b. 1939), Dr. Sc., corresponding member of the Russian Academy of Sciences, graduated from Lomonosov Moscow State University in 1961. Currently he works as head of sector 9 at the Institute for Systems Analysis FRC CSC RAS. Research interests are game theory and pattern recognition. E-mail: *vladimir.arlazarov@smartengines.com* .